\documentclass[10pt,twocolumn,letterpaper]{article}

\usepackage{iccv}
\usepackage{times}
\usepackage{epsfig}
\usepackage{graphicx}
\usepackage{amsmath}
\usepackage{amssymb}
\usepackage{animate}
\usepackage{tabularx}
\usepackage{threeparttable}
\usepackage{booktabs}
\usepackage{multirow} 
\usepackage{balance}
\usepackage{appendix}
\usepackage{listings}
\usepackage{color}

\usepackage[pagebackref=true,breaklinks=true,letterpaper=true,colorlinks,bookmarks=false]{hyperref}

\iccvfinalcopy 

\def\iccvPaperID{****} 

\ificcvfinal\fi

\begin{document}

\title{AdaAttN: Revisit Attention Mechanism in Arbitrary Neural Style Transfer}

\author{Songhua Liu$^{1,2,*}$, Tianwei Lin$^{1,\dag}$, Dongliang He$^1$, Fu Li$^1$, Meiling Wang$^1$,\\Xin Li$^1$, Zhengxing Sun$^{2,\dag}$, Qian Li$^3$, Errui Ding$^1$\\
$^1$Department of Computer Vision Technology (VIS), Baidu Inc.,\\
$^2$Nanjing University, $^3$National University of Defense Technology\\
{\tt\small $^1$\{liusonghua,lintianwei01,hedongliang01,lifu,wangmeiling03,lixin41,dingerrui\}@baidu.com},\\
{\tt\small $^2$songhua.liu@smail.nju.edu.cn, szx@nju.edu.cn, $^3$liqian@nudt.edu.cn}\\
}

\maketitle

\renewcommand{\thefootnote}{\fnsymbol{footnote}}
\footnotetext[1]{This work was done when Songhua Liu was an intern at VIS, Baidu.}
\footnotetext[2]{Corresponding authors.}
\renewcommand{\thefootnote}{\arabic{footnote}}

\ificcvfinal\thispagestyle{empty}\fi

\begin{abstract}
Fast arbitrary neural style transfer has attracted widespread attention from academic, industrial and art communities due to its flexibility in enabling various applications. 
Existing solutions either attentively fuse deep style feature into deep content feature without considering feature distributions, or adaptively normalize deep content feature according to the style such that their global statistics are matched. 
Although effective, leaving shallow feature unexplored and without locally considering feature statistics, they are prone to unnatural output with unpleasing local distortions.  
To alleviate this problem, in this paper, we propose a novel attention and normalization module, named \textbf{Ada}ptive \textbf{Att}ention \textbf{N}ormalization (\textbf{AdaAttN}), to adaptively perform attentive normalization on per-point basis. 
Specifically, spatial attention score is learnt from both shallow and deep features of content and style images. Then per-point weighted statistics are calculated by regarding a style feature point as a distribution of attention-weighted output of all style feature points. Finally, the content feature is normalized so that they demonstrate the same local feature statistics as the calculated per-point weighted style feature statistics. 
Besides, a novel local feature loss is derived based on \textit{AdaAttN} to enhance local visual quality. We also extend \textit{AdaAttN} to be ready for video style transfer with slight modifications.
%
Experiments demonstrate that our method achieves state-of-the-art arbitrary image/video style transfer. 
Codes and models are available\footnote{\href{https://github.com/wzmsltw/AdaAttN}{PaddlePaddle Implementation}.}.
\end{abstract}
\vspace{-0.5cm}

\section{Introduction}
Given a content image $I_c$ and a style image $I_s$, artistic style transfer aims at applying style patterns of $I_s$ onto $I_c$ while preserving content structure of $I_c$ simultaneously, which is widely used in computer-aid art generation.
%
%
%
The seminal work of Gatys \etal~\cite{Gatys_2016_CVPR} proposed an image optimization method that iteratively minimizes the joint content and style loss in the feature space of a pre-trained deep neural network.
This time-consuming optimization process has motivated researchers to explore more efficient approaches.
Johnson \etal~\cite{johnson2016perceptual} alternatively considered using a feed-forward network to generate rendered images directly and enabled real time style transfer.
Since the learned model can only work for one specific style, this method and its following works~\cite{wu2018direction, ulyanov2016texture, ulyanov2017improved, li2016precomputed, liu2017depth, wang2017multimodal, jing2018stroke, Kotovenko_2019_ICCV} are categorized to \textit{Per-Style-Per-Model} method~\cite{jing2019neural}.
In the literature, there are \textit{Multiple-Style-Per-Model} solutions~\cite{dumoulin2016learned, chen2017stylebank, li2017diversified, zhang2018multi} and \textit{Arbitrary-Style-Per-Model}~\cite{huang2017arbitrary, chen2016fast, li2018learning, park2019arbitrary, jing2020dynamic, li2017universal, deng2020arbitrary1, deng2020arbitrary2, yao2019attention, sheng2018avatar, wu2020efanet, gu2018arbitrary} methods.
In the latter case, a model can accept any style image as input and produce stylized results in a single forward pass once upon the model is trained.
Therefore, it is the most flexible and attracts increasing attention from academic, industrial, and art communities.

\begin{figure}[!t]
\centering
    \begin{minipage}{.475\textwidth}
        \includegraphics[width=.24\textwidth]{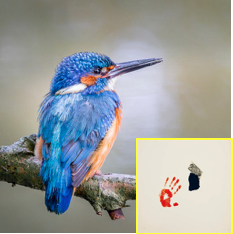}
        \includegraphics[width=.24\textwidth]{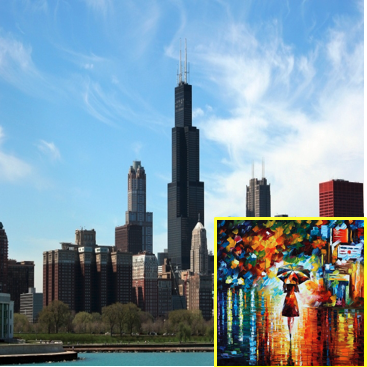}
        \animategraphics[width=.49\textwidth, autoplay, loop]{15}{Figure/Figure_demo/video_c/}{1}{50}
    \end{minipage}\\
    \begin{minipage}{.475\textwidth}
        \includegraphics[width=.24\textwidth]{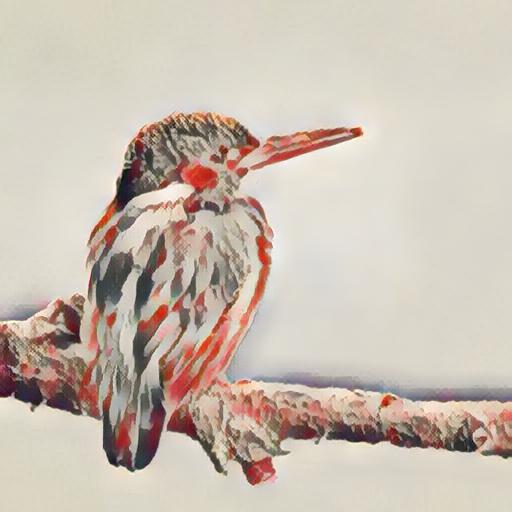}
        \includegraphics[width=.24\textwidth]{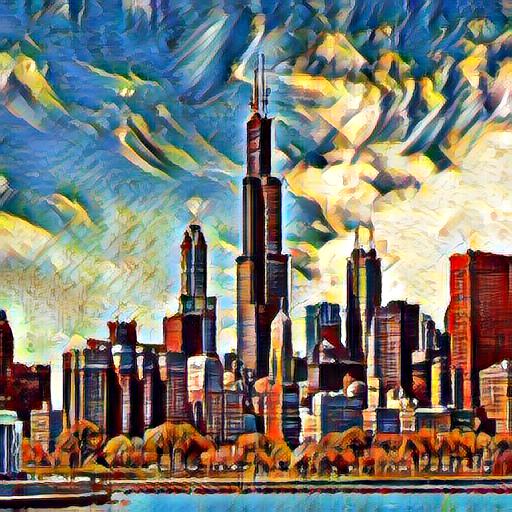}
        \animategraphics[width=.49\textwidth, autoplay, loop]{15}{Figure/Figure_demo/video_cs/}{1}{50}
    \end{minipage}
    \caption{Results generated by our \textit{AdaAttN} methods for arbitrary image/video style. We highly recommend \textit{Adobe Acrobat} to view the \textit{animated clips} at the right side.}
    \label{demo}
    \vspace{-0.5cm}
\end{figure}

Nevertheless, arbitrary style transfer is far from being solved. Enabling flexibility sacrifices local style pattern modeling capability for an arbitrary style transfer network.
For example, the pioneering work \cite{huang2017arbitrary} proposed a simple yet effective method \textit{AdaIN}, which transfers global mean and variance of a style image to a content image in the feature space to support arbitrary input style image.
%
%
Since mean and variance of features are calculated globally, local details and point-wise patterns are largely dismissed and thus the local stylization performance is largely degraded ~\cite{park2019arbitrary}.
Similar trade-off between flexibility and capability also exists in \cite{deng2020arbitrary2, li2018learning, jing2020dynamic, li2017universal, ghiasi2017exploring}, where all local feature points of the content image are processed by the same transformation function based on style images.
%
To enhance the locality awareness of arbitrary style transfer models, recently, attention mechanism is adopted in multiple works \cite{park2019arbitrary, deng2020arbitrary1, yao2019attention} for this task.
Their common intuition is that a model should pay more attention to those feature-similar areas in the style image for stylizing a content image region.
%
%
%
%
Such attention mechanism has been proved to be effective for generating more local style details in arbitrary style transfer.
Unfortunately, while improving the performance, it fails to totally solve this problem and the local distortions still occur. 

It is not so hard to reveal the reasons of the above dilemma posed to the attention mechanism. Digging into the details of current attention based solutions for arbitrary style transfer, it can be easily figured out that 1) the designed attention mechanisms are commonly based on deep CNN features on higher abstraction levels and the low-level details are dismissed; 2) the attention scores are usually used to re-weight feature maps of the style image and the re-weighted style feature is simply fused into content feature for decoding. Deep CNN features based attention strategy leaves the low-level patterns of images at shallow network layers unexplored. Thus the attention scores may focus little on low-level textures and are dominated by high-level semantics. Meanwhile, the spatial re-weighting of style features followed by fusion of re-weighted style features and content features, as done in SANet \cite{park2019arbitrary} (Figure \ref{fig:module}(b)), works without consideration of feature distribution.

To this end, we attempt to address these issues and get a better balance between style pattern transferring and content structure preserving. Motivated by lessons learned through the above analysis, we propose a novel attention and normalization module named \textit{\textbf{Ada}ptive \textbf{Att}ention \textbf{N}ormalization (\textbf{AdaAttN})} for arbitrary style transfer. It can adaptively perform attentive normalization on per-point basis for feature distribution alignment. 
In more detail, spatial attention score is learnt from both shallow and deep features of content and style images.
Then, per-point weighted statistics is calculated by regarding a style feature point as a distribution of attention-weighted output of all spatial feature points. 
Finally the content feature is normalized so that its local feature statistics are the same as the per-point weighted style feature statistics. 
In this way, the attention module takes into account both shallow and deep CNN features of the style image as well as the content image. Meanwhile, alignment of per-point feature statistics from the content feature to the modulated style feature is achieved.
Based on \textit{AdaAttN} module, a novel optimization objective named \textit{local feature loss} and a new arbitrary image style transfer pipeline are derived.
Our contributions can be summarized as follows:
\begin{itemize}
\item We introduce a novel \textit{AdaAttN} module for arbitrary style transfer. It takes both shallow and deep features into account for attention score calculation and properly normalizes content feature such that feature statistics are well aligned with attention-weighted mean and variance maps of style features on per-point basis.
\item A new optimization objective called \textit{local feature loss} is proposed. It helps the model training and improves arbitrary style transfer quality by regularizing local features of the generated image.
\item Extensive experiments and comparisons with other state-of-the-art methods are performed to demonstrate the effectiveness of our proposed method. 
\item Further extension of our model for video style transfer via simply introducing cosine-distance based attention and image-wise similarity loss can result in stable and appealing results.
\end{itemize}


\section{Related Works}


\subsection{Arbitrary Style Transfer}
Recent arbitrary style transfer methods can be divided into two categories: global transformation based and local transformation based methods. 
The common idea of the former category is to apply feature modification globally. WCT~\cite{li2017universal} achieved this with two transformation steps including whitening and coloring.
Huang \etal~\cite{huang2017arbitrary} proposed AdaIN that adaptively applies mean and standard deviation of each style feature to shift and re-scale the corresponding normalized content feature so that content feature and style feature share the same distribution.
Jing \etal~\cite{jing2020dynamic} extended this method by dynamic instance normalization, where weights for intermediate convolution blocks are generated by another network taking the style image as input.
Li \etal~\cite{li2018learning} proposed to generate a linear transformation according to content and style features.
Furthermore, Deng \etal~\cite{deng2020arbitrary2} got the transformation function with multi-channel correlation. 
Although these methods accomplish the overall arbitrary style transfer task and make great progress in this field, local style transfer performance is generally unsatisfactory due to global transformations leveraged by them are hard to take care of detailed local information.

For the latter, Chen \etal~\cite{chen2016fast} proposed a style-swap method, which is a patch based style transfer method relying on similarities between content and style patches. 
\cite{gu2018arbitrary} was another patch based method considering matching of both global statistics and local patches. 
Avatar-Net~\cite{sheng2018avatar} further proposed a multi-scale framework that combines ideas of style swap and AdaIN function.
In recent years, attention mechanism is widely used in arbitrary style transfer thanks to its excellent ability to model fine grained correspondence among local features of style and content images. 
On this routine, Park \etal~\cite{park2019arbitrary} proposed Style-Attentional Network (\textit{SANet}) to match content and style features.
Yao \etal~\cite{yao2019attention} considered different types of strokes with such attention framework.
Deng \etal~\cite{deng2020arbitrary1} proposed a multi-adaptation module that applies point-wise attention for content features and channel-wise attention for style features.
Common practices adopted by these methods are to build the attention mechanism merely upon deep CNN features without considering shallow features and simply mix the content feature and the re-weighted style feature. 
Thus, it tends to distort original content structures largely and results in undesired effect for human eyes. 
In this paper, our goal is to explore a better trade-off between style pattern transferring and content structure preserving. 

\subsection{Video Style Transfer}
Directly applying image style transfer techniques on video frame sequences usually results in flickering effects caused by temporal inconsistency. 
Thus, a lot of works add optical flow consistency constraint to original image style transfer solutions, \eg, \cite{ruder2016artistic} for optimization based video style transfer, \cite{ruder2018artistic, chen2017coherent, gupta2017characterizing, huang2017real, gao2018reconet} for per-style-per-model methods, \cite{wang2020consistent, ReReVST2020} for arbitrary-style-per-model methods, and \cite{wang2018video, chen2019mocycle, liu2020stable} for image-to-image translation frameworks.
Optical flow constraint improves the stability of video style transfer.
However, it heavily relies on a pre-extracted optical flow field with high accuracy to perform flow-based warping.
%
There are also some works that address the stability issue with approaches other than optical flow warping.
\cite{li2018learning, deng2020arbitrary2} leveraged linearity of transformation models to guarantee inter-frame consistency on feature space.
Wu \etal~\cite{wu2020preserving} proposed a \textit{SANet} based method that leads current frame to focus on similar regions of previous frame with the help of a \textit{SSIM} consistency constraint. 
Different from these methods, in this work, we add a novel image-wise similarity loss based on attention mechanism to overcome the flickering artifact and comparable or even better stability is achieved without prerequisite optical flow.

\section{Methods}

\begin{figure}
\begin{center}
\includegraphics[width=\linewidth]{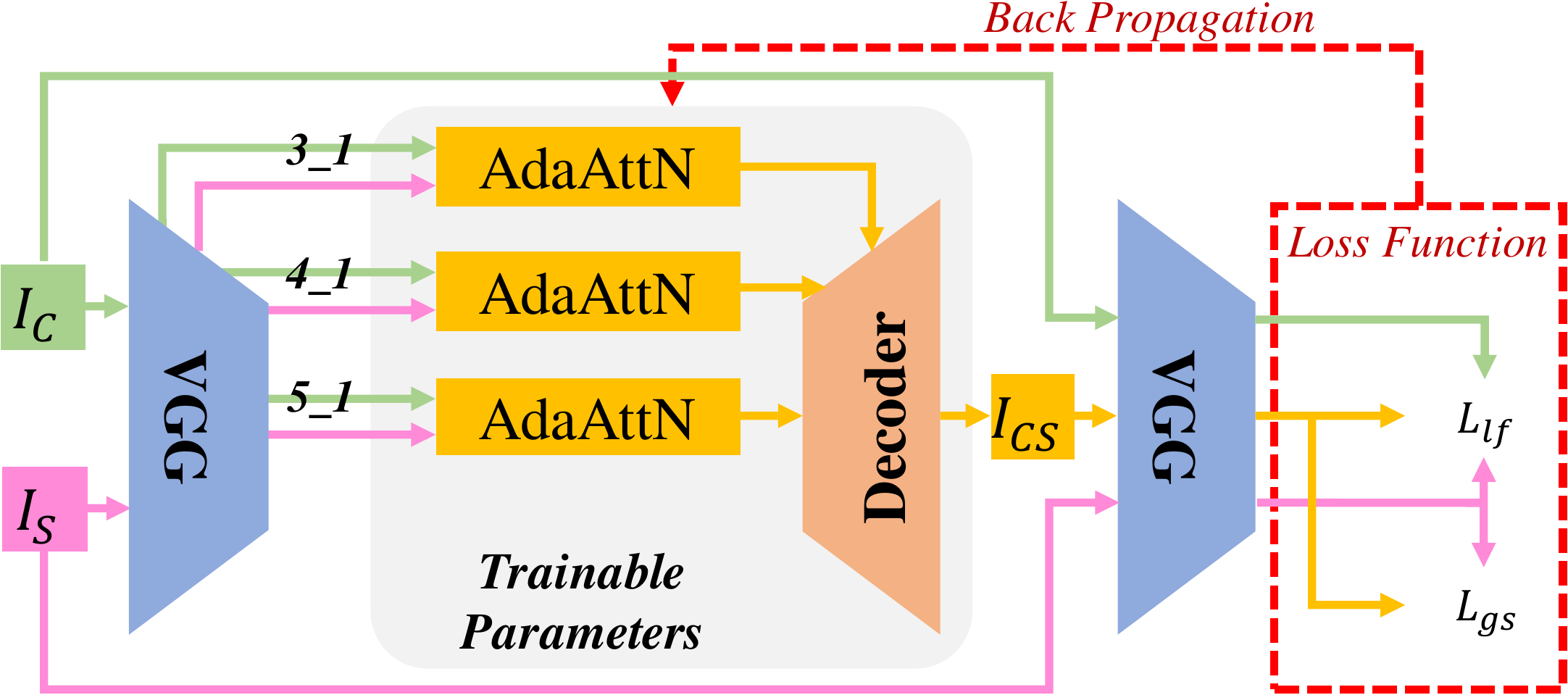}
\end{center}
   \caption{Overview of full framework, where three \emph{AdaAttN} modules and the decoder are trainable. $\mathcal{L}_{lf}$ and $\mathcal{L}_{gs}$ are local feature loss and global style loss separately.}
\label{fig_overallframework}
\end{figure}

\begin{figure*}
\begin{center}
\includegraphics[width=.9\linewidth]{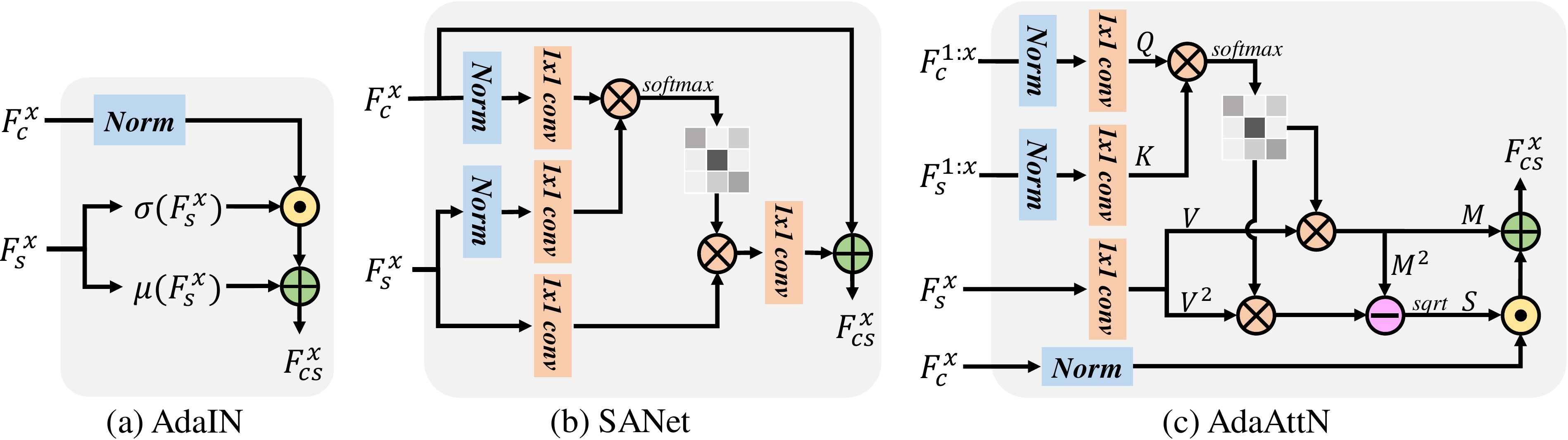}
\end{center}
   \caption{(a) The structure of \textit{AdaIN} \cite{huang2017arbitrary} module; (b) The structure of \textit{SANet} \cite{park2019arbitrary} module; (c) The structure of our proposed \textit{AdaAttN} module. $Norm$ here denotes the mean-variance channel-wise normalization.}
\label{fig:module}
\end{figure*}


\subsection{Overall Framework}

The proposed network takes a style image $I_s$ and a content image $I_c$ to synthesize a stylized image $I_{cs}$.
In our proposed model, we employ a pre-trained VGG-19 network \cite{simonyan2015very} as encoder to extract multi-scale feature maps. The decoder follows the setting of \cite{huang2017arbitrary} with a symmetric structure of VGG-19. 
In order to take full advantage of features in both shallow and deep levels, we employ a multi-level strategy by integrating three AdaAttN modules on \textit{ReLU-3\_1} , \textit{ReLU-4\_1} and \textit{ReLU-5\_1} layers  of VGG, respectively, as shown in Figure \ref{fig_overallframework}.
We denote the extracted feature of layer \textit{ReLU-x\_1} in VGG as $F_*^x\in R^{C\times H_*W_*}$ when it takes an image $I_*$ as input and $*$ can be $c$ or $s$ here representing content and style features respectively. 
To fully exploit low-level patterns, we further concatenate feature of current layer with down-sampled features of its previous layers as:

\begin{equation}
F_*^{1:x}=D_x(F_*^{1})\oplus D_x(F_*^{2})\oplus\cdots\oplus F_*^x,
\end{equation}
where $D_x$ stands for the bilinear interpolation layer which downsamples the input feature to the same shape of $F_*^x$, and $\oplus$ here means concatenation operation along channel dimension. Then, we can denote the embedded feature of the AdaAttN module at layer $l$ as:

\begin{equation}
F^x_{cs}=AdaAttN(F_c^x, F_s^x, F_c^{1:x}, F_s^{1:x}),
\end{equation}
where $F_c$, $F_s$ and $F_{cs}$ are content, style, and embedded feature, respectively. With multi-level embedded features, we can synthesize the stylized image $I_{cs}$ with decoder as:
\begin{equation}
I_{cs} = Dec(F^3_{cs}, F^4_{cs}, F^5_{cs}).
\end{equation}


\subsection{Adaptive Attention Normalization}

The feature transformation module is the key component in arbitrary style transfer models. A comparison of our module with other frameworks is demonstrated in Figure \ref{fig:module}.
The pioneering AdaIN \cite{huang2017arbitrary} only considers the holistic style distributions and the content feature is manipulated such that its feature distribution globally aligns with that of the style feature.
By taking local style patterns into consideration, SANet \cite{park2019arbitrary} calculates attention map from the style and content feature maps and then modulates the style feature with the attention map to fuse the attention output into the content feature. SANet performs in local stylization. However, it lacks low-level matching and local feature distribution alignment.
Inspired by lessons learned from AdaIN and SANet, we propose the \emph{Adaptive Attention Normalization (AdaAttN)} module, which can adaptively transfer feature distribution on per-point basis via considering both low-level and high-level features with attention mechanism.
As shown in Figure \ref{fig:module}(c), AdaAttN works in three steps: \textbf{(1)} computing attention map with content and style features from shallow to deep layers; \textbf{(2)} calculating weighted mean and standard variance maps of style feature; \textbf{(3)} adaptively normalizing content feature for per-point feature distribution alignment.

\textbf{Attention Map Generation.}
In arbitrary style transfer methods, attention mechanism is used to measure the similarity between content and style features.
Different from previous methods which only use relatively deep features, we use low-level and high-level layers of both content and style feature simultaneously.
%
To computing attention map $A$ of layer $x$, we formulate $Q$ (query), $K$ (key) and $V$ (value) as:
\begin{equation}
\begin{aligned}
    Q&=f(Norm(F_c^{1:x})),\\
    K&=g(Norm(F_s^{1:x})),\\
    V&=h(F_s^x),\\
\end{aligned}
\end{equation}
where $f$, $g$, and $h$ are $1\times 1$ learnable convolution layers, $Norm$ here denotes channel-wise mean-variance normalization, as used in instance normalization. 
The attention map $A$ can be calculated as:

\begin{equation}
    A = Softmax(Q^{\top}\otimes K),\label{eq_attn}
\end{equation}
where $\otimes$ denotes matrix multiplication.

\textbf{Weighted Mean and Standard Variance Map.}
Applying attention score matrix $A$ to style feature $F_s^x$ as SANet \cite{park2019arbitrary} does can be regarded as calculating each target style feature point by weighted summation of all style feature points. 
In this paper, we interpret this process as viewing a target style feature point by attention output as a distribution of all the weighted style feature points. Then from this perspective, we can calculate statistics for each distribution. We term such statistics as \textit{attention-weighted mean} and \textit{attention-weighted standard variance} respectively. Thus, the attention-weighted mean $M\in R^{C\times H_cW_c}$ becomes:
\begin{equation}
    M=V\otimes A^{\top},
\end{equation}
%
where $A \in R^{H_cW_c\times H_sW_s}$ and $V\in R^{C\times H_sW_s}$.
Since variance of a variable equals to the expectation of its square minus the square of its expectation, we can obtain the attention-weighted standard deviation $S\in R^{C\times H_cW_c}$ as:
\begin{equation}
    S=\sqrt{(V\cdot V) \otimes A^{\top} - M\cdot M},
\end{equation}
where $\cdot$ denotes element-wise product. 

\textbf{Adaptive Normalization.}
Finally, for each position and each channel of normalized content feature map, corresponding scale in $S$ and shift in $M$ are used to generate transformed feature map:

\begin{equation}
F_{cs}^x=S\cdot Norm(F_c^x)+M.
\end{equation}

In short, AdaAttN performs feature statistics transferring via generating attention-weighted mean and variance maps. As shown in Figure \ref{fig:module}, compared with AdaIN, AdaAttN considers per-point statistics rather than globally. AdaAttN is more general than AdaIN. For each $i,j$,  if set $A_{i,j}=1/(H_sW_s)$, AdaAttN can be specialized to AdaIN. Compared with SANet, attention mechanism is used to calculate target feature distribution instead of directly generating transferred feature for further fusion. 
%

\begin{figure*}
\begin{center}
\includegraphics[width=.975\linewidth]{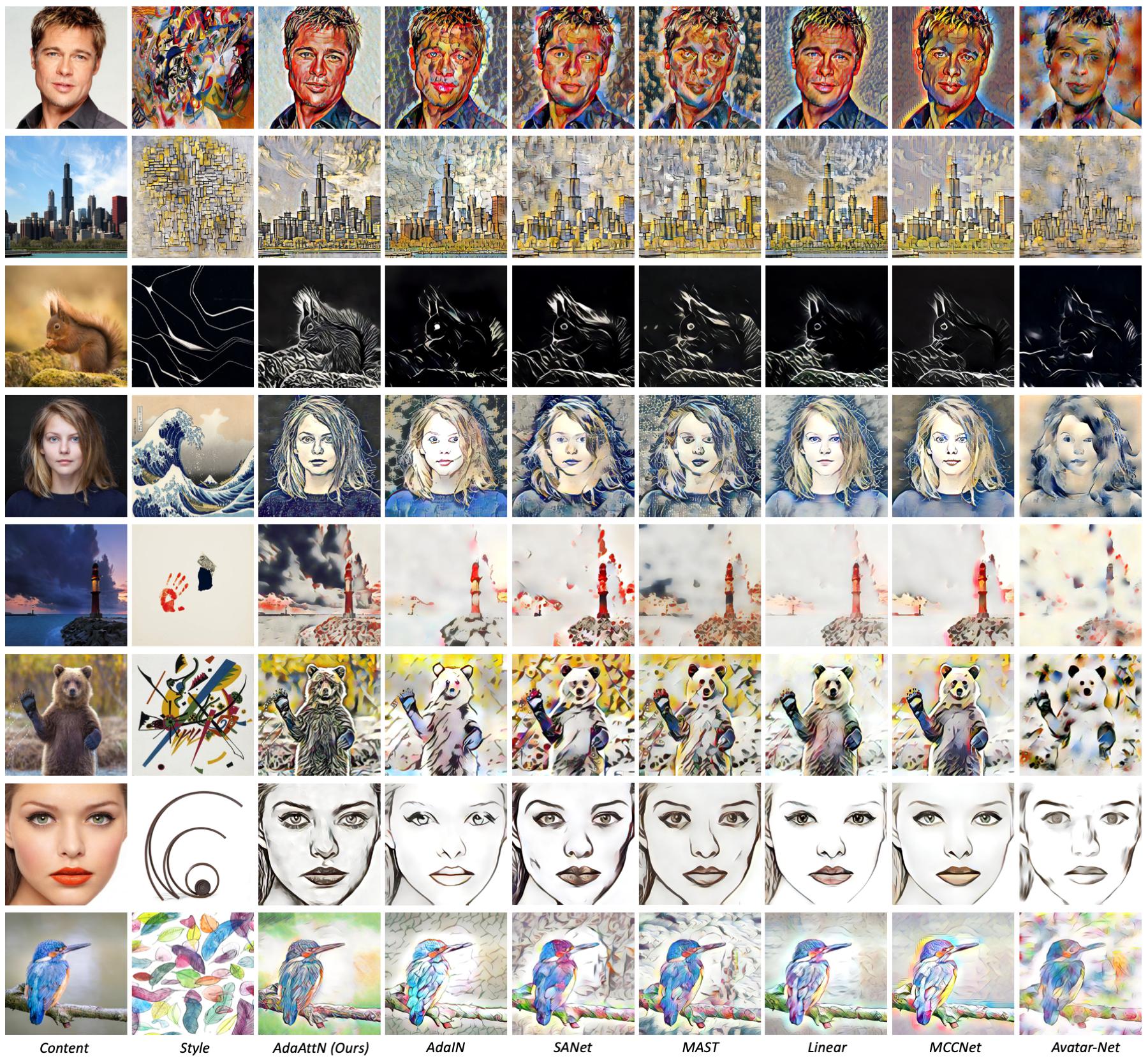}
\end{center}
\caption{Comparison with other state-of-the-art methods in arbitrary image style transfer.}
\label{fig:compare_sota}
\end{figure*}

\begin{figure*}
\begin{center}
\includegraphics[width=.975\linewidth]{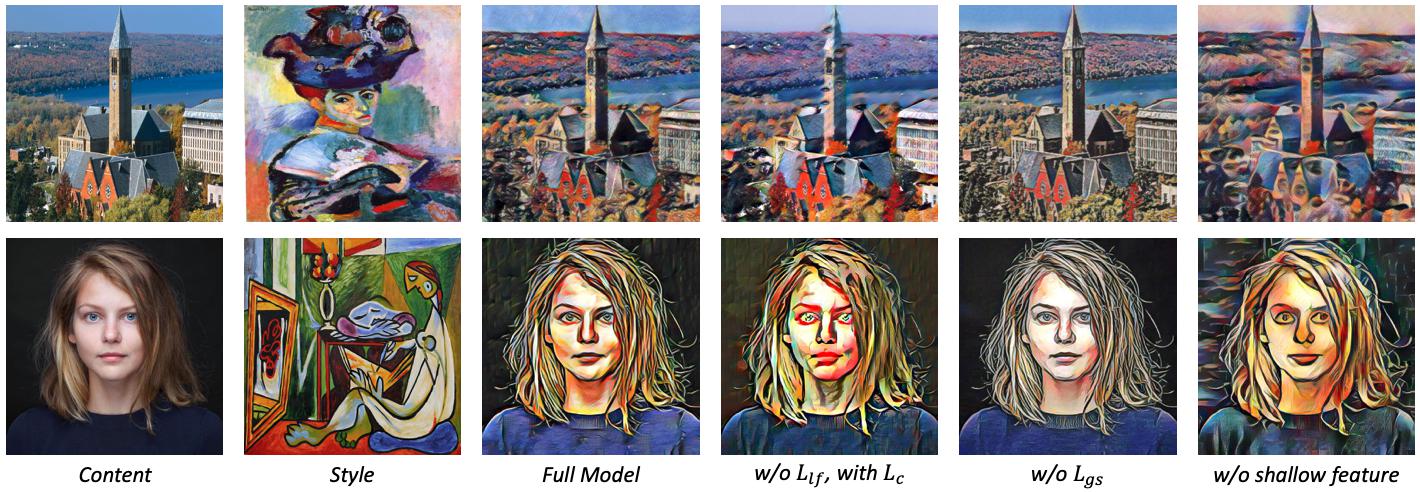}
\end{center}
\vspace{-0.2cm}
\caption{Ablation study on loss functions and shallow feature. Zoom-in for better view.}
\vspace{-0.3cm}
\label{fig:ablation}
\end{figure*}

\subsection{Loss Function}
Our overall loss function is the weighted summation of global style loss ($\mathcal{L}_{gs}$) and local feature loss ($\mathcal{L}_{lf}$):
\begin{equation}
    \mathcal{L}=\lambda_g\mathcal{L}_{gs}+\lambda_l\mathcal{L}_{lf},
\end{equation}
where $\lambda_g$ and $\lambda_l$ are hyper-parameters controlling weights of their corresponding loss terms.
Details of each loss term will be explained in the remaining part of this section.

To begin with, following \cite{huang2017arbitrary} and many other works, distances of mean $\mu$ and standard deviation $\sigma$ between generated image and style image in VGG feature space are penalized to guarantee global stylized effect ($\mathcal{L}_{gs}$):
\begin{equation}
\begin{aligned}
    \mathcal{L}_{gs}=\sum_{x=2}^5(\left|\left|\mu(E^x(I_{cs}))-\mu(F^x_s)\right|\right|_2\\+\left|\left|\sigma(E^x(I_{cs}))-\sigma(F^x_s)\right|\right|_2),
\end{aligned}
\end{equation}
where $E()$ denotes feature of the VGG encoder and its superscript $^x$ denotes the layer index.

The proposed novel local feature loss $\mathcal{L}_{lf}$ constrains that feature map of stylized image is consistent with the transformation result by \textit{AdaAttN} function:
\begin{equation}
    \mathcal{L}_{lf}=\sum_{x=3}^5\left|\left|E^x(I_{cs})-AdaAttN^*(F_c^x, F_s^x, F_c^{1:x}, F_s^{1:x})\right|\right|_2, 
\end{equation}
where $AdaAttN^*$ serves as a supervision signal that should be deterministic. Thus, we consider the parameter-free version of $AdaAttN$ without the three learnable $1\times 1$ convolution kernels ($f$, $g$, and $h$).
Local feature loss makes the model generates better stylized output for local areas compared with conventional content loss term used in \cite{huang2017arbitrary,park2019arbitrary}.

\subsection{Extension for Video Style Transfer}
Compared to other attention-based methods, our method is capable of generating more natural stylized results without much local distortions, and thus it is of great potential for video style transfer. With two slight modifications, our model can be extended to video style transfer.
%
%
\begin{figure}[t]
\begin{center}
\includegraphics[width=\linewidth]{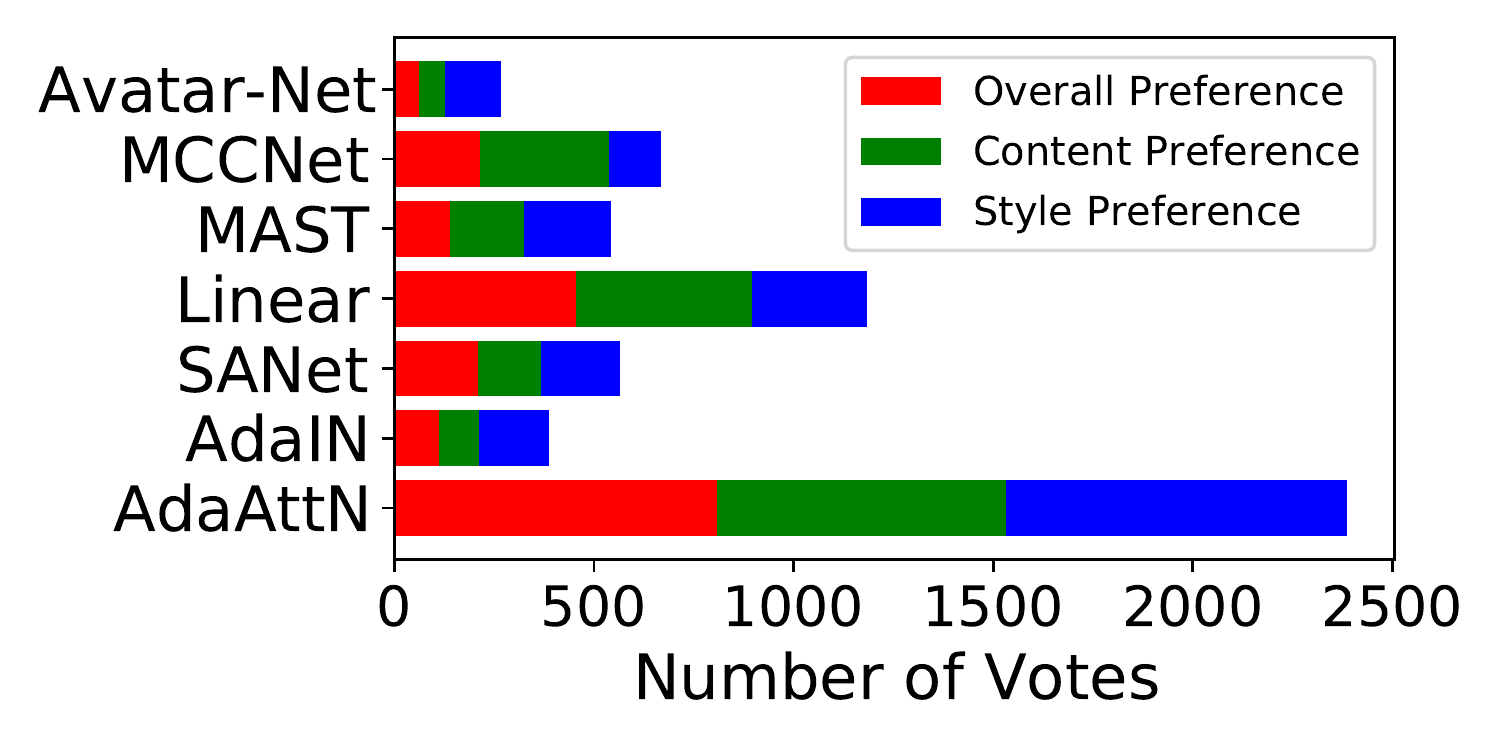}
\end{center}
\vspace{-0.5cm}
\caption{Results of user study.}
\vspace{-0.5cm}
\label{fig:user_study}
\end{figure}

First of all, we notice that \textit{Softmax} function in Eq.\ref{eq_attn} shows strong exclusiveness in attention score due to exponential computation and it can focus majorly on local patterns and has negative influence in stabilization. For video style transfer,
alternatively, we consider cosine similarity for attention map computation:
\begin{equation}
    A_{i,j}=\frac{S_{i,j}}{\sum_{j}S_{i,j}},
    S_{i,j}=\frac{Q_i\cdot K_j}{\left|\left|Q_i\right|\right|\times\left|\left|K_j\right|\right|}+1,
\label{att_cos}
\end{equation}
where cosine similarity results in more flat attention score distribution than Softmax. 
Thus, the local feature statistics will be more stable and local style patterns will not be over emphasised, leading to better assurance of consistency.

Secondly, based on attention mechanism, we design a novel cross-image similarity loss term $\mathcal{L}_{is}$ to regularize the relevant contents between two content images $c_1,c_2$:
\begin{equation}
\begin{aligned}
    \mathcal{L}_{is}=\sum_{x=2}^4\frac{1}{N_{c_1}^{x}N_{c_2}^{x}}&\sum_{i,j}\left|\frac{D_{c_1,c_2}^{i,j,x}}{\sum_{i}D_{c_1,c_2}^{i,j,x}}-\frac{D_{cs_1,cs_2}^{i,j,x}}{\sum_{i}D_{cs_1,cs_2}^{i,j,x}}\right|,\\
    D_{u,v}^{i,j,x}=1&-\frac{F_u^{x,i}\cdot F_v^{x,j}}{\left|\left|F_u^{x,i}\right|\right|\times\left|\left|F_v^{x,j}\right|\right|},
\end{aligned}
\end{equation}
where $N_c^x$ is the size of spatial dimension in content feature map $F_c^x$ of layer \textit{ReLU-x\_1}, $F_*^{x,i}$ means feature vector of $i$-th position of $F_{*}^x$, and $D_{u,v}^{i,j,x}$ measures cosine distance of $F_u^{x,i}$ and $F_v^{x,j}$.
In each training iteration, two input video frames are sampled to enable this loss.
Intuitively, such cross-image similarity loss requires the stylized results of two content images share similar local similarity patterns with the two original images. Therefore, it ensures awareness of inter-frame relationship in video style transfer and contributes to stable results.

\begin{table}[!t]
\centering
    \begin{tabular}{lccc}
        \toprule
        \multirow{2}{1.5cm}{Method} & \multicolumn{3}{c}{Inference Time (sec./image)} \\
        \cline{2-4}
        & $256\times256$ & $512\times256$ & $512\times512$ \\
        \midrule
        Avatar-Net & 0.124 & 0.176 & 0.311\\
        AdaIN & 0.038 & 0.049 & 0.066\\
        Linear & 0.028 & 0.036 & 0.049\\
        MCCNet & 0.024 & 0.040 & 0.057\\
        MAST & 0.046 & 0.073 & 0.115\\
        SANet & 0.043 & 0.064 & 0.081\\
        \hline
        Ours & 0.051 & 0.066 & 0.112 \\
        \bottomrule
    \end{tabular}
    \caption{Running speed comparison under different resolutions.}
    \label{table:efficiency}
    \vspace{-0.5cm}
\end{table}

\begin{figure*}[!t]
\centering
    \begin{minipage}{\textwidth}
    \centering
        \includegraphics[width=.24\textwidth]{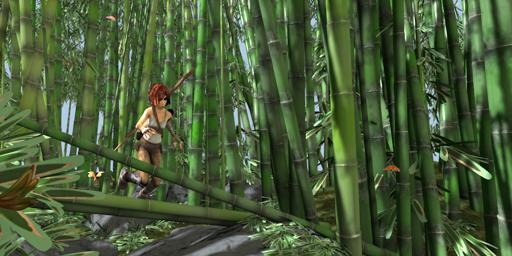}
        \includegraphics[width=.24\textwidth]{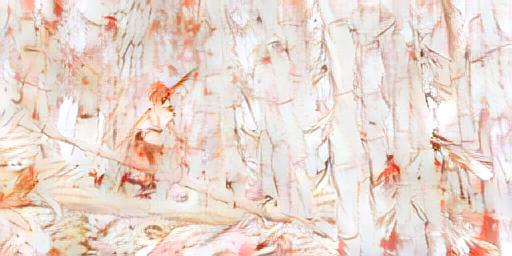}
        \includegraphics[width=.24\textwidth]{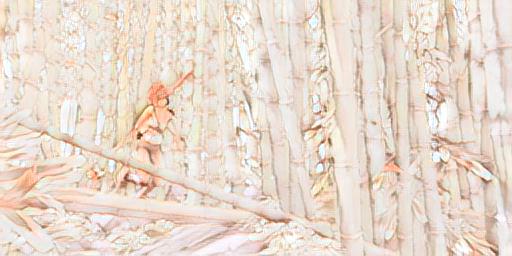}
        \includegraphics[width=.24\textwidth]{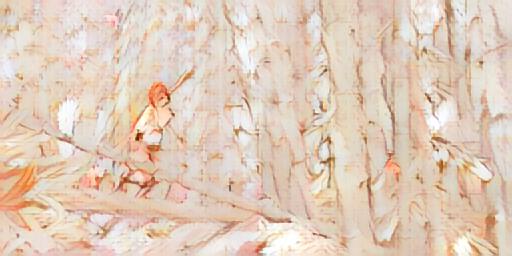}
    \end{minipage}\\
    \begin{minipage}{\textwidth}
    \centering
        \begin{minipage}{.24\textwidth}
        \centering
            \includegraphics[width=.5\textwidth]{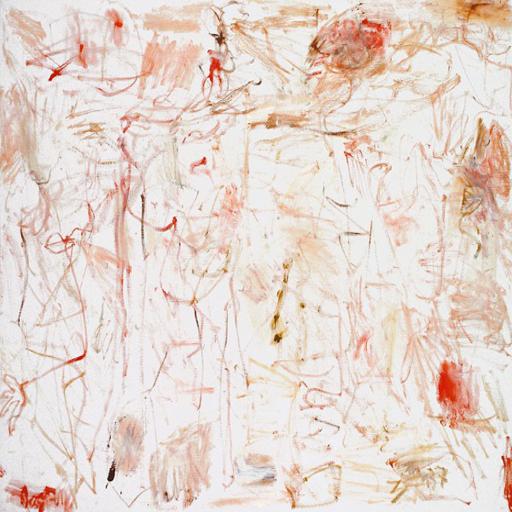}
            \centerline{Content \& Style}
        \end{minipage}
        \begin{minipage}{.24\textwidth}
        \centering
            \includegraphics[width=\textwidth]{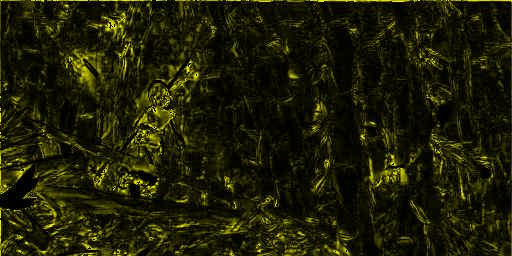}
            \centerline{SANet}
        \end{minipage}
        \begin{minipage}{.24\textwidth}
        \centering
            \includegraphics[width=\textwidth]{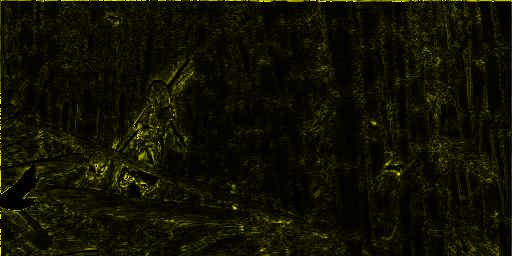}
            \centerline{Linear}
        \end{minipage}
        \begin{minipage}{.24\textwidth}
        \centering
            \includegraphics[width=\textwidth]{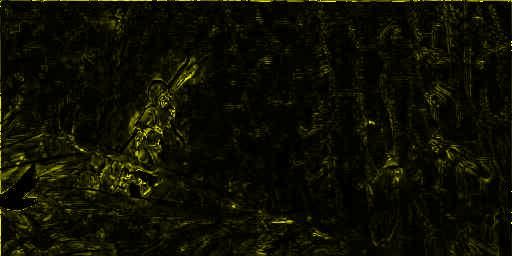}
            \centerline{MCCNet}
        \end{minipage}
    \end{minipage}\\
    \begin{minipage}{\textwidth}
    \centering
        \includegraphics[width=.24\textwidth]{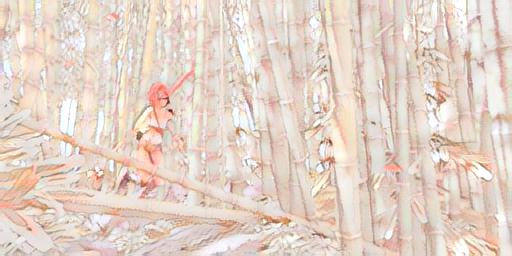}
        \includegraphics[width=.24\textwidth]{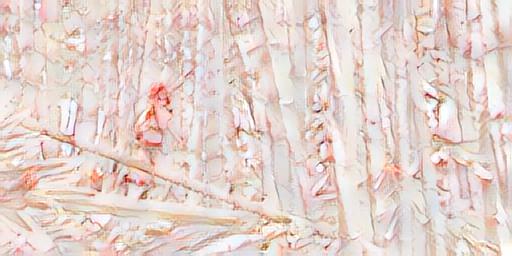}
        \includegraphics[width=.24\textwidth]{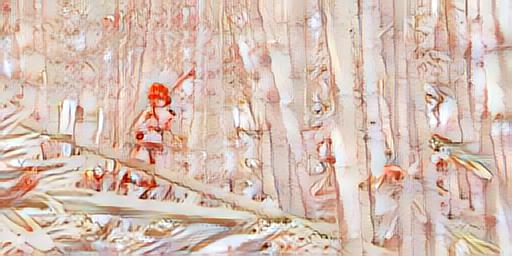}
        \includegraphics[width=.24\textwidth]{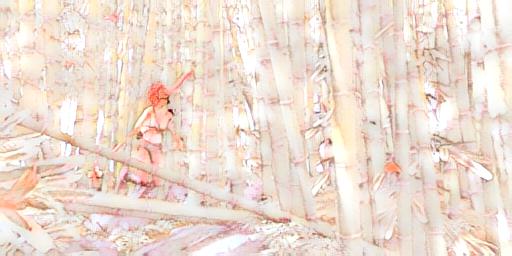}
    \end{minipage}\\
    \begin{minipage}{\textwidth}
    \centering
        
        \begin{minipage}{.24\textwidth}
        \centering
            \includegraphics[width=\textwidth]{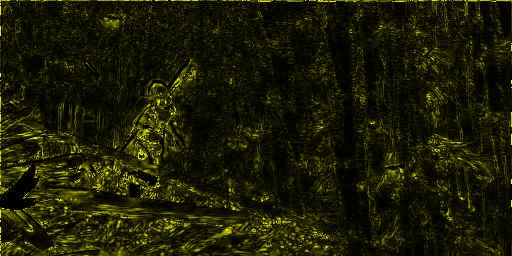}
            \centerline{Ours}
        \end{minipage}
        \begin{minipage}{.24\textwidth}
        \centering
            \includegraphics[width=\textwidth]{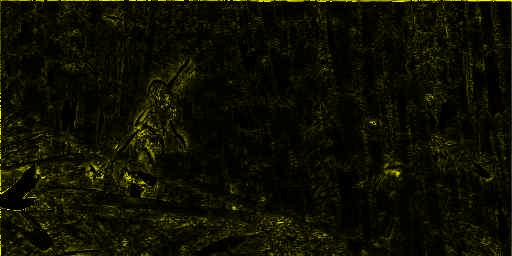}
            \centerline{Ours + $Cos$}
        \end{minipage}
        \begin{minipage}{.24\textwidth}
        \centering
            \includegraphics[width=\textwidth]{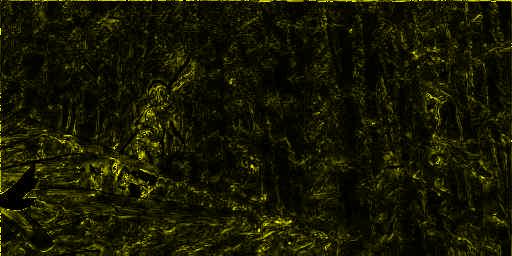}
            \centerline{Ours + $\mathcal{L}_{is}$}
        \end{minipage}
        \begin{minipage}{.24\textwidth}
        \centering
            \includegraphics[width=\textwidth]{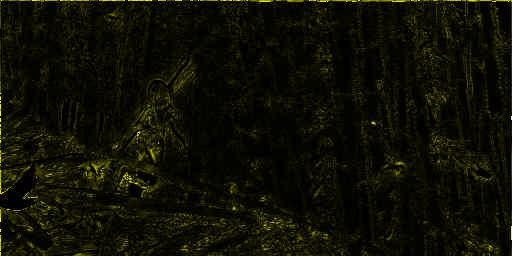}
            \centerline{Ours + $Cos$ + $\mathcal{L}_{is}$}
        \end{minipage}
    \end{minipage}\\
    \caption{Qualitative comparison among different methods or settings on video style transfer. The first row shows results by different methods or settings. The second row shows the corresponding optical flow error map.}
    \label{fig:compare_video}
    \vspace{-0.5cm}
\end{figure*}

\section{Experiments}
\subsection{Implementing Details}
We train our arbitrary style transfer model with \textit{MS-COCO}~\cite{lin2014microsoft} as our content image set and \textit{WikiArt}~\cite{phillips2011wiki} as our style image set.
%
$\lambda_{g}$, $\lambda_{l}$, and $\lambda_{is}$ (for video style transfer only) are set as $10$, $3$, and $100$, respectively.
\textit{Adam}~\cite{kingma2014adam} with $\alpha$, $\beta_1$, and $\beta_2$ of $0.0001$, $0.9$, and $0.999$, is used as solver.
In the training phase, all images are loaded with $512\times 512$ resolution and randomly cropped to $256\times 256$ for augmentation.
While inference, our model can be applied to images with any resolution.
In this section, $512\times 512$ and $512\times 256$ resolutions are used for image and video, respectively.
The training lasts for $50$K iterations on a single Nvidia Tesla P40 GPU and batch size is 8 for image and 4 for video.
Please refer to appendix for detailed network configurations.

\begin{figure*}[t]
\begin{center}
\includegraphics[width=0.95\linewidth]{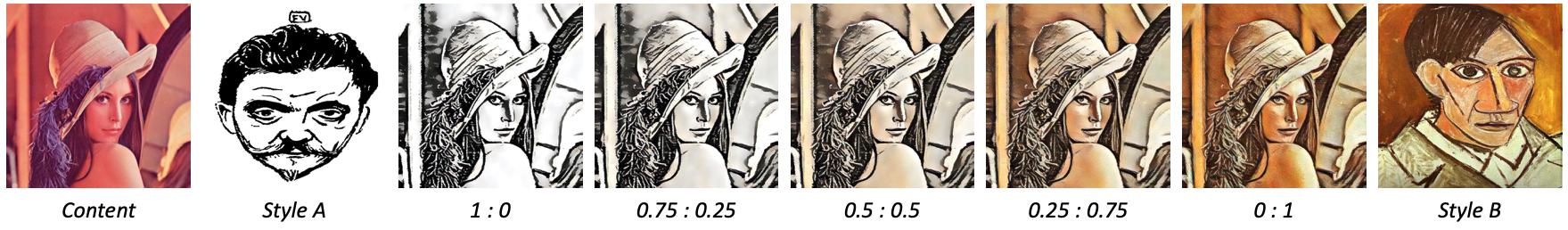}
\end{center}
\vspace{-0.5cm}
\caption{Style interpolation.}
\vspace{-0.5cm}
\label{fig_interp}
\end{figure*}

\subsection{Comparison with State-of-the-Art Methods}

\noindent
\textbf{Qualitative Comparison.}
As shown in Figure \ref{fig:compare_sota}, we compare our method with six state-of-the-art arbitrary style transfer methods, including AdaIN \cite{huang2017arbitrary}, SANet \cite{park2019arbitrary}, MAST \cite{deng2020arbitrary2}, Linear \cite{li2018learning}, MCCNet \cite{deng2020arbitrary2}, and Avatar-Net \cite{sheng2018avatar}.
AdaIN \cite{huang2017arbitrary} directly adjusts second-order statistics of content feature globally and we can see style patterns are transferred with severe content details lost (1st, 5th and 6th rows).
Avatar-Net \cite{sheng2018avatar} utilizes AdaIN for multi-scale transferring and introduces style decorators with patch matching strategies, which results in blur stylization results with obvious plaques (1st, 6th and 8th rows).
SANet \cite{park2019arbitrary} and MAST \cite{deng2020arbitrary2} adopt attention mechanism to attentively transfer style features to content features in deep layers. It will result in damaged content structures (3rd, 4th and 6th rows) and dirty textures (1st, 2nd, 8th rows). Some style patches are even directly transferred into the content image improperly (4th, 8th rows).
Linear \cite{li2018learning} and MCCNet \cite{deng2020arbitrary2} modify features via linear projection and per-channel correlation respectively, both resulting in relative clean stylization outputs. However, textural patterns of style images are not captured adaptively, loss of content details is encountered (3rd, 5th and 6th rows), and content image color remains (7th row).
As shown in the 3rd column, AdaAttN can adaptively transfer style patterns to each location of content images appropriately, attributing to the novel attentive normalization on per-point.
It shows that AdaAttN achieves a better balance between style transferring and content structure preserving.

\noindent
\textbf{User Study.}
Following SANet, 15 content images and 15 style images are randomly picked to form 225 images pairs in total. Then we randomly sample 20 content-style pairs and synthesize stylized images by different methods. Results are presented side-by-side in a random order and we ask subjects to select their favorite one from three views: content preservation, stylization degree, and overall preference. 
We collect 2000 votes for each view from 100 users and show the number of votes of each method in the form of bar chart.
The results in Figure \ref{fig:user_study} demonstrate that our stylized results are more appealing than competitors.

\noindent
\textbf{Efficiency Analysis.}
We demonstrate run time performance of AdaAttN and SOTA feed-forward methods on Table \ref{table:efficiency}. All experiments are conducted using a single Nvidia P40 GPU.
Although multi-depth feature layers (from $1\_1$ to $5\_1$) are used, our method can still achieve 20 FPS at 256px, which is comparable with SOTA methods such as SANet \cite{park2019arbitrary} and Linear \cite{li2018learning}. 
Thus, our proposed AdaAttN can practicably synthesize stylized images in real time.

\subsection{Ablation Study}

\begin{table}[!t] 
	\begin{tabular}{m{2.6cm}<{\centering}m{0.6cm}<{\centering}m{0.6cm}<{\centering}m{0.6cm}<{\centering}m{0.6cm}<{\centering}m{0.7cm}<{\centering}}
	\toprule
  		Method & Style1 & Style2 & Style3 & Style4 & Mean \\ 
  	\hline 
		SANet & 8.57 & 8.93 & 10.3 & 4.66 & 7.76 \\ 
		Linear & 4.41 & 5.10 & 5.24 & 2.67 & 4.42 \\ 
		MCCNet & 4.63 & 4.84 & 5.48 & 2.35 & 4.45 \\ 
	\hline
	Ours  & 5.65 & 5.77 & 6.41 & 3.39 & 5.52 \\
	Ours + Cos & 4.09 & 4.59 & 5.15 & 2.26 & 4.09 \\ 
	Ours + $\mathcal{L}_{is}$ & 5.51 & 5.31 & 6.26 & 3.31 & 5.51 \\ 
	Ours + Cos + $\mathcal{L}_{is}$ & \textbf{3.70} & \textbf{4.46} & \textbf{4.49} & \textbf{2.14} & \textbf{3.91} \\
  	\bottomrule
 	\end{tabular}
 	\caption{Optical flow error ($\times10^{-2}$) of SOTA methods and different AdaAttN variants. Smaller values mean better temporal consistency. $Cos$ here stands for attention score of cosine similarity. The mean value is calculated using 20 styles.}
 	\label{table:video}
 	\vspace{-0.5cm}
\end{table}

\noindent
\textbf{Loss Function.}
As shown in Figure \ref{fig:ablation}, we present ablation study results to verify the effectiveness of each loss term used for training AdaAttN.
\textbf{(1)} To verify the effectiveness of our proposed local feature loss $\mathcal{L}_{lf}$, we replace it with the vanilla \emph{L2} content loss $\mathcal{L}_c$ which constraints feature distance between $I_c$ and $I_{cs}$ and is used in many style transfer methods \cite{huang2017arbitrary, deng2020arbitrary1, park2019arbitrary}. As shown in the 4th column, their visual quality is obviously worse than that of the full model. 
It suggests that compared to content loss, our proposed local feature loss can better take style patterns into consideration while preserving content structures.
\textbf{(2)} We remove the global style loss $\mathcal{L}_{gs}$ and train model with $\mathcal{L}_{lf}$ only. As shown in the 5th column, style patterns are also weakly transferred without style loss, which demonstrates that $\mathcal{L}_{lf}$ can force network to learn style transfer to some extent. However, the overall color saturation is degraded, showing that the global style loss is necessary. 

\noindent
\textbf{Low-level Feature.}
To verify the effectiveness of shallow feature used in AdaAttN, we remove shallow feature via replacing the $Q$ and $K$ of AdaAttN from $F^{1:x}$ to $F^x$. Some local content damage and dirty textures can be observed (last column of Figure \ref{fig:ablation}). Our AdaAttN can effectively utilize shallow features to generate pleasant stylization results.

\subsection{Video Style Transfer}

For video stylization, we compare our method with SOTA methods SANet, Linear and MCCNet, where optical flow is not used for stabilization. To verify the effectiveness of our proposed methods for video stylization, we also provide ablation results of adding $Cos$ and $\mathcal{L}_{is}$, where $Cos$ denotes attention score of cosine similarity (Eq.\ref{att_cos}).
The qualitative results in Figure \ref{fig:compare_video} and quantitative results in Table. \ref{table:video} both demonstrate that (1) our method is more stable than attention-based method \emph{SANet}; (2) replacing Softmax activation with cosine distance based attention can significantly improve temporal consistency; (3) with our proposed modifications, \emph{AdaAttN} is more stable than \emph{Linear} and \emph{MCCNet}, which are proposed for video stylization.

\subsection{Multi-Style Transfer}

\begin{figure}[t]
\begin{center}
\includegraphics[width=0.80\linewidth]{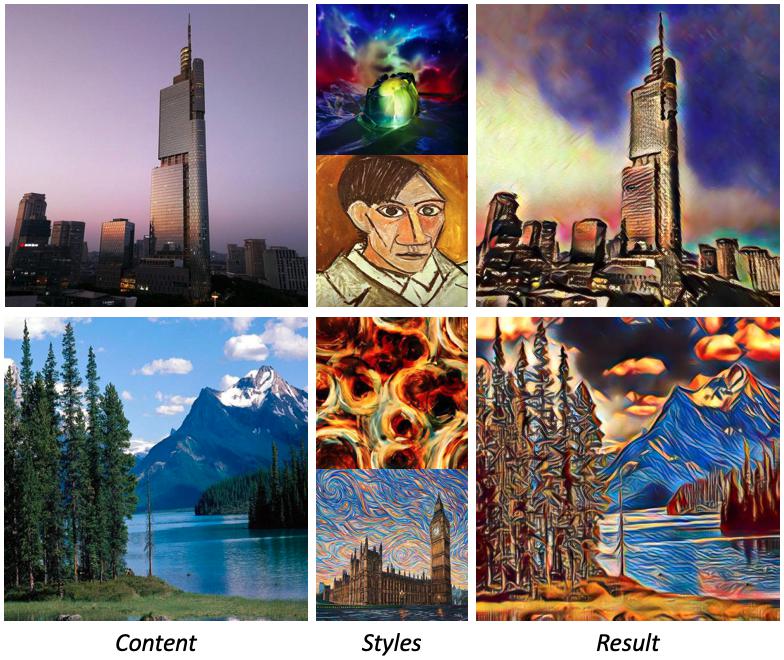}
\end{center}
\vspace{-0.5cm}
\caption{Results of multi-style transfer.}
\vspace{-0.5cm}
\label{fig:multi_style}
\end{figure}

\noindent
Following previous works \cite{park2019arbitrary,deng2020arbitrary1}, we explore interpolating several style images via averaging their mean and standard variance maps of different styles, then the combined mean and variance are used to modulate the content feature for decoding (Figure \ref{fig_interp}).
Besides style interpolation, we can also achieve multi-style transfer via concatenating multiple style images into one image and feeding it into AdaAttN (Figure \ref{fig:multi_style}). From these results, we can see AdaAttN can flexibly support various run-time controls with plausible outcomes.

\section{Conclusion}
In this paper, we propose a novel \textit{AdaAttN} module for arbitrary style transfer. AdaAttN performs feature statistics transferring via modulation with per-point attention-weighted mean and variance of style feature. Attention weights are generated from both style and content features from low-level to high-level. With slight modifications, our model is ready for video style transfer as well.
Experiment results demonstrate that our method can generate high quality stylization results for both images and videos. AdaAttN has the potential for improving other image manipulation or translation tasks, we will explore this in our future work.

%
%

{\small
\bibliographystyle{ieee_fullname}
\balance
\bibliography{AttnAdaIN}
}

\newpage
\appendix
\section{Network Details}

\begin{table}[!t]
\centering
\begin{tabular}{c|c|c}
\hline
 Stage            & Output            &   Architecture \\ \hline
$F^5$ & $512\times\frac{H}{8}\times\frac{W}{8}$ & \begin{tabular}{c} Input $F^5_{cs}$ \\
Upsample, scale 2 \\
Add $F^4_{cs}$ \\
$3\times3$ Conv, 512, ReLU \end{tabular} \\ \hline
$F^4$    & $256\times\frac{H}{4}\times\frac{W}{4}$  &   \begin{tabular}{c} $3\times3$ Conv, 256, ReLU \\
Upsample, scale 2 \end{tabular} \\ \hline
$F^3$     & $128\times\frac{H}{2}\times\frac{W}{2}$ &   \begin{tabular}{c} Concatenate $F^3_{cs}$ \\
($3\times3$ Conv, 256, ReLU)$\times3$ \\
$3\times3$ Conv, 128, ReLU \\
Upsample, scale 2 \end{tabular} \\ \hline
$F^2$     & $64\times H \times W$  &   \begin{tabular}{c} $3\times3$ Conv, 128, ReLU \\
$3\times3$ Conv, 64, ReLU \\
Upsample, scale 2 \end{tabular} \\ \hline
$F^1$     & $3\times H \times W$  &   \begin{tabular}{c} $3\times3$ Conv, 64, ReLU \\
$3\times3$ Conv, 3 \end{tabular} \\ \hline
\end{tabular}
\vspace{0.2cm}
\caption{Architecture of our decoder network.}
\label{table_network}
\end{table}

\subsection{Decoder}
The decoder of our framework takes results of three AdaAttN modules on \textit{ReLU-3\_1}, \textit{ReLU-4\_1}, and \textit{ReLU-5\_1} layers as input.
Similar to the decoder of SANet, feature on \textit{ReLU-5\_1} is upsampled to the same size as that of \textit{ReLU-4\_1}, followed by element-wise addition.
Then, there is a learnable $3\times 3$ convolution block for feature transformation.
The following architecture is symmetrical with VGG encoder (up to \textit{ReLU-4\_1}), except that the number of input channels is twice on \textit{ReLU-3\_1} layer in order to incorporate AdaAttN output on this level.
Full decoder configuration is shown in Table \ref{table_network}.

\subsection{AdaAttN}

We provide \textit{PyTorch} code of AdaAttN module.
The implementation is elegant and its overall time and space complexities are the same as SANet~\cite{park2019arbitrary}.

\definecolor{dkgreen}{rgb}{0,0.6,0}
\definecolor{gray}{rgb}{0.5,0.5,0.5}
\definecolor{mauve}{rgb}{0.58,0,0.82}

\lstset{frame=tb,
  language=Python,
  aboveskip=3mm,
  belowskip=3mm,
  showstringspaces=false,
  columns=flexible,
  basicstyle={\small\ttfamily},
  numbers=none,
  numberstyle=\tiny\color{gray},
  keywordstyle=\color{blue},
  commentstyle=\color{dkgreen},
  stringstyle=\color{mauve},
  breaklines=false,
  breakatwhitespace=true,
  tabsize=3
}

\begin{lstlisting}
class AdaAttN(nn.Module):
    def __init__(self, v_dim, qk_dim):
        super().__init__()
        self.f = nn.Conv2d(qk_dim, qk_dim, 1)
        self.g = nn.Conv2d(qk_dim, qk_dim, 1)
        self.h = nn.Conv2d(v_dim, v_dim, 1)

    def forward(self, c_x, s_x, c_1x, s_1x):
        Q = self.f(mean_variance_norm(c_1x))
        Q = Q.flatten(-2, -1).transpose(1, 2)
        K = self.g(mean_variance_norm(s_1x))
        K = K.flatten(-2, -1)
        V = self.h(s_x)
        V = V.flatten(-2, -1).transpose(1, 2)
        A = torch.softmax(torch.bmm(Q, K), -1)
        M = torch.bmm(A, V)
        Var = torch.bmm(A, V ** 2) - M ** 2
        S = torch.sqrt(Var.clamp(min=0))
        M = M.transpose(1, 2).view(c_x.size())
        S = S.transpose(1, 2).view(c_x.size())
        return S * mean_variance_norm(c_x) + M
\end{lstlisting}

\begin{figure}[t]
\begin{center}
\includegraphics[width=\linewidth]{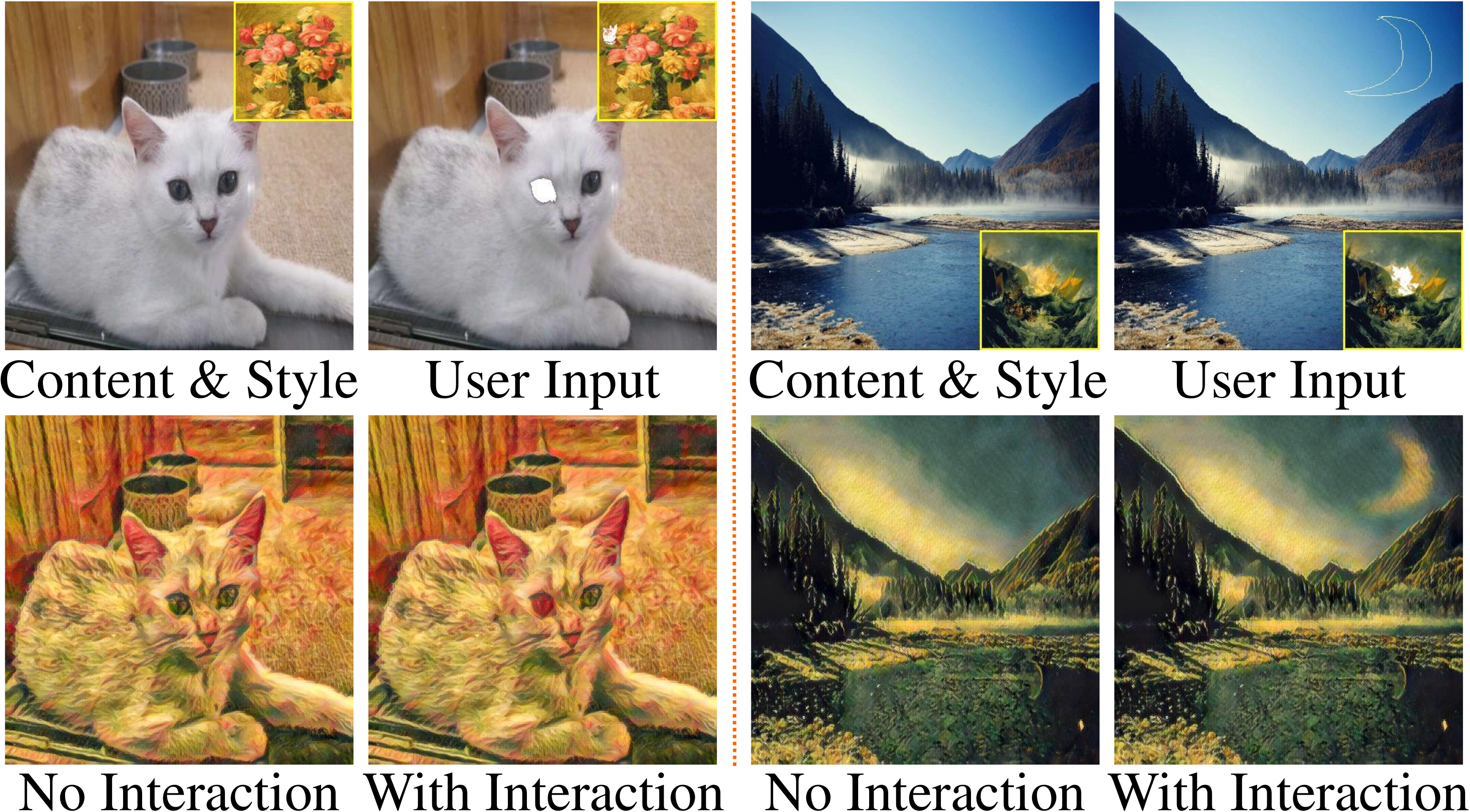}
\end{center}
\caption{User controlling demos.}
\label{fig_user}
\end{figure}

\begin{figure*}[t]
\begin{center}
\includegraphics[width=\linewidth]{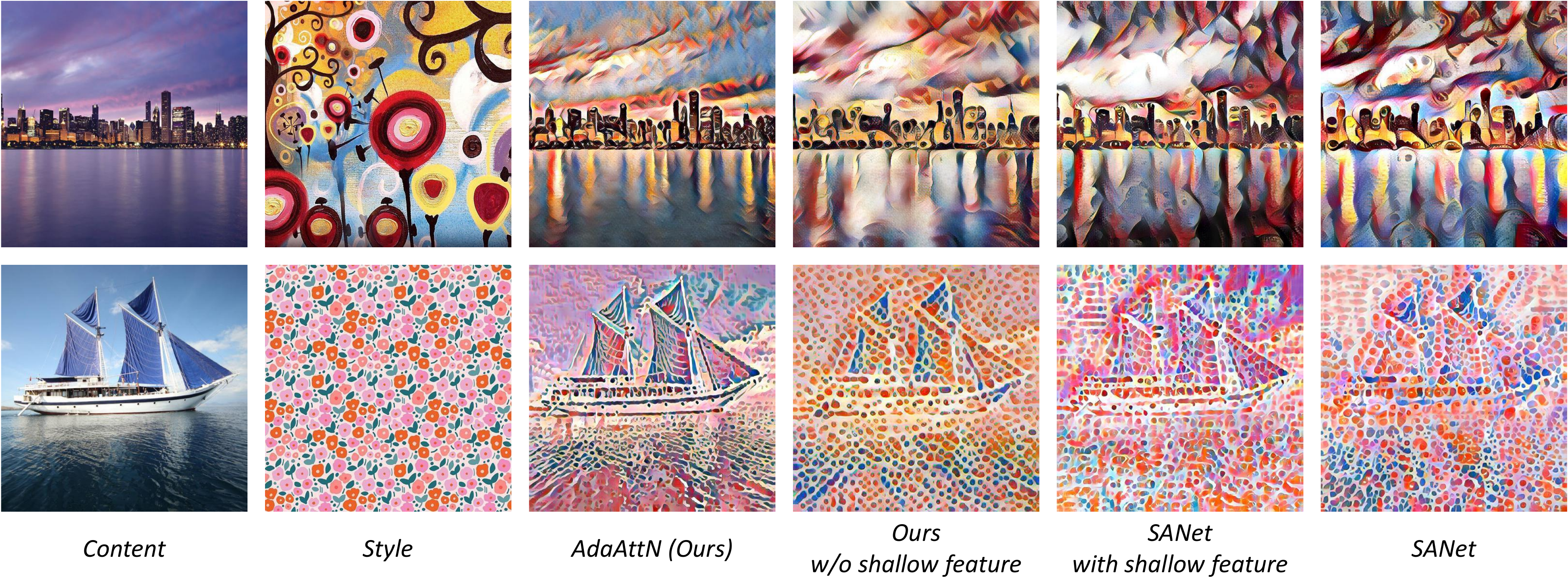}
\end{center}
\caption{More ablation study results.}
\label{fig_ablation}
\end{figure*}

\begin{table*}[!t] 
\centering
	\begin{tabular}{m{1.7cm}<{\centering}m{0.33cm}<{\centering}m{0.33cm}<{\centering}m{0.33cm}<{\centering}m{0.33cm}<{\centering}m{0.33cm}<{\centering}m{0.33cm}<{\centering}m{0.33cm}<{\centering}m{0.33cm}<{\centering}m{0.33cm}<{\centering}m{0.33cm}<{\centering}m{0.33cm}<{\centering}m{0.33cm}<{\centering}m{0.33cm}<{\centering}m{0.33cm}<{\centering}m{0.33cm}<{\centering}m{0.33cm}<{\centering}m{0.33cm}<{\centering}m{0.33cm}<{\centering}m{0.33cm}<{\centering}m{0.33cm}}
	\toprule
  		Method & 1 & 2 & 3 & 4 & 5 & 6 & 7 & 8 & 9 & 10 & 11 & 12 & 13 & 14 & 15 & 16 & 17 & 18 & 19 & 20 \\ 
  	\hline 
		SANet & 8.57 & 8.93 & 10.3 & 4.66 & 12.4 & 4.39 & 9.06 & 5.31 & 10.6 & 11.5 & 12.0 & 3.29 & 8.92 & 5.82 & 7.58 & 8.40 & 5.84 & 4.97 & 4.51 & 8.21 \\ 
		Linear & 4.41 & 5.10 & 5.24 & 2.67 & 6.73 & 2.68 & 6.69 & 2.19 & 5.03 & 7.80 & 7.50 & 1.90 & 4.84 & 3.69 & 4.59 & 4.35 & 2.75 & 3.13 & 2.98 & 4.03 \\ 
		MCCNet & 4.63 & 4.84 & 5.48 & 2.35 & 6.92 & 2.39 & 8.26 & 2.72 & 5.75 & \textbf{6.70} & 7.34 & 1.93 & 4.16 & 3.64 & 4.47 & 4.25 & 3.05 & 2.94 & 2.84 & 4.29 \\ 
	\hline
    	Ours  & 5.65 & 5.77 & 6.41 & 3.39 & 8.36 & 4.00 & 7.08 & 4.78 & 6.73 & 8.76 & 8.48 & 2.61 & 6.16 & 4.38 & 5.55 & 6.00 & 3.55 & 3.75 & 3.55 & 5.37  \\
    	Ours + Cos & 4.09 & 4.59 & 5.15 & 2.26 & 6.59 & \textbf{2.24} & \textbf{5.97} & \textbf{2.06} & 4.89 & 7.45 & 7.27 & 1.70 & 4.43 & 3.35 & 4.06 & \textbf{3.94} & 2.53 & 2.78 & 2.68 & 3.69 \\ 
    	Ours + $\mathcal{L}_{is}$ & 5.51 & 5.31 & 6.26 & 3.31 & 7.96 & 4.56 & 6.84 & 5.03 & 6.37 & 8.90 & 8.55 & 2.62 & 6.15 & 4.82 & 5.30 & 6.29 & 3.66 & 4.01 & 3.63 & 5.05 \\ 
    	Ours + Cos + $\mathcal{L}_{is}$ & \textbf{3.70} & \textbf{4.46} & \textbf{4.49} & \textbf{2.14} & \textbf{6.06} & 2.52 & 6.24 & 2.15 & \textbf{4.55} & 7.35 & \textbf{7.11} & \textbf{1.60} & \textbf{3.86} & \textbf{3.27} & \textbf{3.85} & 4.03 & \textbf{2.31} & \textbf{2.54} & \textbf{2.44} & \textbf{3.43} \\
  	\bottomrule
 	\end{tabular}
 	\vspace{0.2cm}
 	\caption{Full results of optical flow error evaluation for 20 styles.}
 	\label{table_video_more}
\end{table*}

\newpage
\nobalance

\section{More Results}
\subsection{Image Style Transfer}
\noindent\textbf{User Control.} Our method can support user-controlled stylization conveniently.
User-specified content regions would adopt features of user-specified style regions by manipulating attention map used in AdaAttN module.
In practice, user can either choose points on content and style images by mouse click (\textit{e.g.}, Figure \ref{fig_user} (left)), or outlining regions with closed borders (\textit{e.g.}, Figure \ref{fig_user} (right)).
Then, user-specified regions for content and style images can be generated by means of the classical region growing algorithm.
To achieve use-controlled stylization, simply setting the attention scores between specified content regions and out of interest style regions as $-\infty$ before the Softmax operation in AdaAttN can work very well.

\noindent\textbf{More Ablation.} As discussed in our main paper, there are two factors leading to distorted stlylization of SANet: absence of low-level feature and failure in distribution alignment. 
To further illustrate impacts of these factors, we conduct more ablation studies under four settings: (1) AdaAttN, (2) AdaAttN without shallow features, (3) SANet with shallow features, and (4) SANet.
As shown in Figure \ref{fig_ablation}, both shallow features and feature distribution alignment prevent dirty textures to some extent.
Combining them together, AdaAtN in this paper receives the best stylization results with the least distortion.

\begin{figure*}
\begin{center}
\includegraphics[width=.975\linewidth]{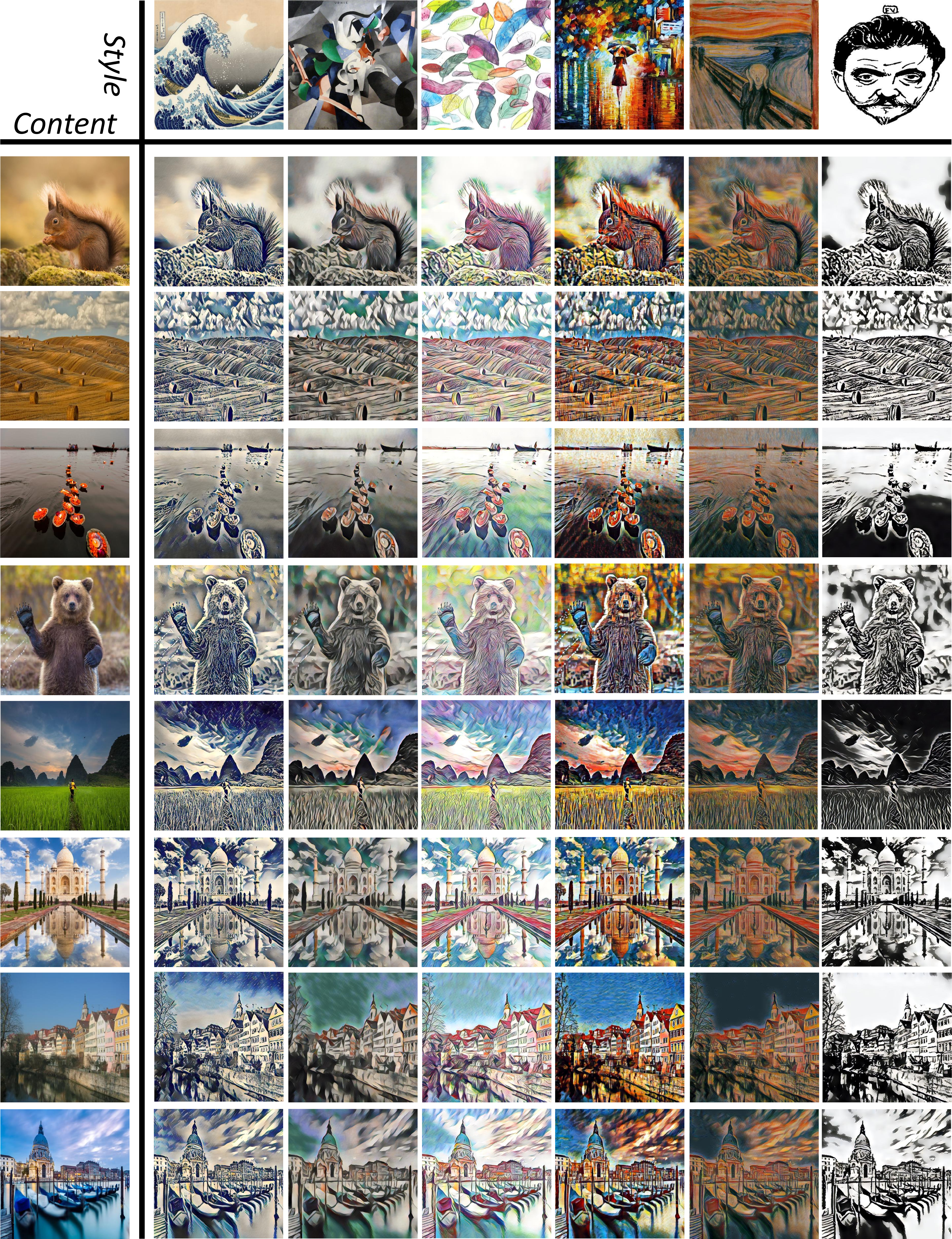}
\end{center}
   \caption{More image style transfer results.}
\label{fig_img}
\end{figure*}

\noindent\textbf{Pair-wise Combination between Content and Style.} In order to demonstrate the robustness of our method on different contents and styles, we provide stylization results of pair-wise combinations between 8 content images and 6 style images in Figure \ref{fig_img}. It can be seen that our AdaAttN can robustly achieve appealing style transfer results. 

\subsection{Video Style Transfer}
\noindent\textbf{Quantitative Results.} Optical flow errors of all the 20 styles\footnote{They are from \hyperref[https://github.com/xunhuang1995/AdaIN-style]{official codebase of AdaIN}.} used for video stylization are shown in Table \ref{table_video_more}, as supplement to Table \ref{table:video}.

\begin{figure*}
\begin{center}
\includegraphics[width=.975\linewidth]{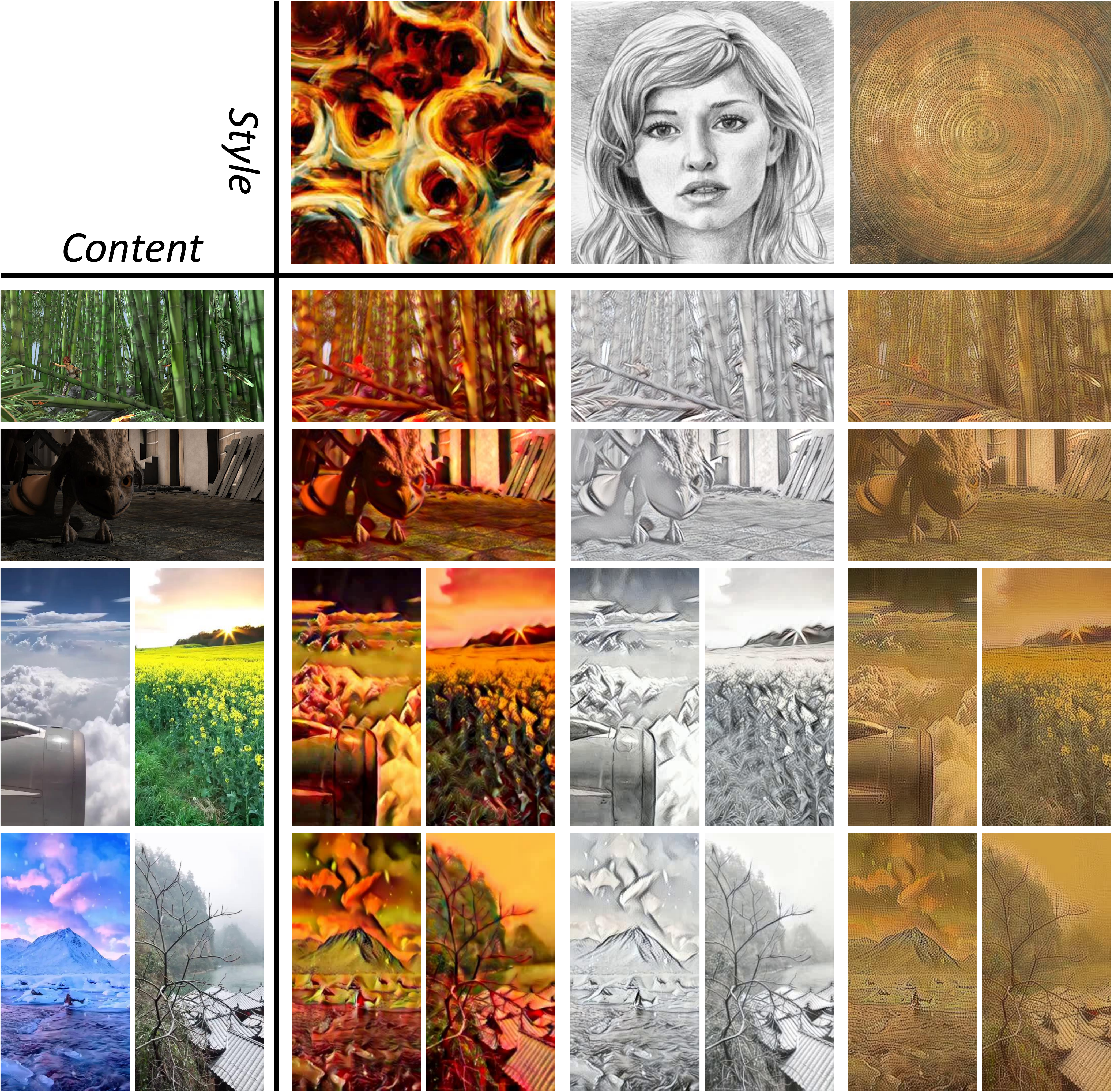}
\end{center}
   \caption{Snapshots of video stylization results. Full videos can be found in the attachments.}
\label{fig_video}
\end{figure*}

\noindent\textbf{Qualitative Results.} We provide more video style transfer examples in Figure \ref{fig_video}. Full animations can be found in the attachments.

\end{document}